**PackingGPT: 3D Packing Agent for Real Furniture in Last-Mile Delivery**

**Yi You***
Faculty of Technology, Policy and Management
Technische Universiteit Delft, Delft, Netherlands
Email: y.yi@tudelft.nl

**Hui Li**
Department of Industrial and Systems Engineering
The Hong Kong Polytechnic University, Hong Kong, China
Email: hui5li@polyu.edu.hk

## ABSTRACT

**Objectives:** 3D bin packing rectangular items into standardised containers to maximise space utilisation under geometric shipping automation. Loading a furniture purchase into a personal vehicle is the same task, but under more complex conditions that standard container loading algorithms ignore. This paper addresses the physically stable placement under these realistic conditions with heterogeneous boxes (e.g. varying dimensions and weights) and occupied containers (e.g. groceries).

**Methods**: This paper provides a real-world benchmark dataset and baseline model for the Heterogeneous furniture-in-vehicle packing task. The dataset uses real furniture company flat-pack packaging data covering a large number of catalogue products via family-level extrapolation with diversity length, widths, heights, and weights. We also propose a Lego-Language Packing (LLP) framework for packing as a sequential placement inspired by the Lego assembly process, where heterogeneous boxes of varying dimensions (bricks) are placed step-by-step into the irregular remaining cargo space (creations).

**Findings**: Five baseline packing methods were tested on our dataset without considering the Centre-of-Mass (CoM) constraints. In sedan car simulations, 10-40% of placed boxes failed the stability check on average. When the LLP model was trained on packing sequences with CoM constraints enforced during placement, the failure rate dropped to 0.67% (SUV-500). It is worth noticing that our LTP model is generalizable, where the eval cases are from unseen products placed in entirely different customer cargo configurations without additional fine-tuning.

**Novelty**: The work first considers the physical balance problem under a real-world scenario, which enables extension to broader logistics planning, directly supporting last-mile delivery route planning and customer self-pickup.

**Practical Applications:** The model supports pre-trip packing feasibility checks given a customer order and an available vehicle, predicting whether the items will fit and remain stable in transit.

## INTRODUCTION

Loading 3D items into a confined space is one of the most common yet computationally intractable problems in logistics, such as packing containers at sail ports or packaging items for delivery. (Ramos 2016)When it comes to the furniture scenario, millions of IKEA customers face the same question every day: will the flat-pack boxes fit in their personal car? This heterogeneous packing problem is defined as: given a set of flat-pack boxes with known packaging dimensions and weights, how to determine the packing steps where the boxes should be put into a daily **cluttered trunk** with **stability constraints**.

Researchers have been focusing on the 3D packing problem for a long time since heuristic algorithms enforce constraints (Bortfeldt 2013) on geometry to *maximise volume utilisation*. Though the influence of gravity has drawn growing attention from society (Ma 2025), it is only considered after placements where the cargo is treated as a whole for transportation. The flat-pack furniture packing problem dramatically amplifies the impact of balance for each box; on one hand, most flat-pack boxes are thin and tall for volume compression; on the other hand, the real cargo truck is rarely empty, with irregularity staffs such as strollers, groceries, and sports equipment already occupying it.

Not only the balance constraint, but the scenario is complex and full of randomness, such as the randomness of the *number and category of the products*, the fill percentage and *the layout of the cargo* trunk. This challenges rule-based heuristic algorithms. Although recent deep learning methods (Yang 2024) learn effective policies for complex situations (Que 2023), they share a common limitation where they are trained for homogeneous bin sizes and do not generalise across heterogeneous vehicle geometries. This leads to an urgent need for a large packing model. (Zhao 2022)

In this paper, we propose a real-world benchmark dataset and baseline model for the heterogeneous furniture-in-vehicle packing (HFVP) task. (Hassamontr 2003) The dataset uses real furniture company flat-pack packaging data covering a large number of catalogue products via family-level extrapolation with diversity length, widths, heights, and weights. We also propose a Lego-Language Packing (LLP) framework for packing as a sequential placement inspired by the Lego assembly process (Pun 2025), where heterogeneous boxes of varying dimensions (bricks) are placed step-by-step into the irregular remaining cargo space (creations).

**(1) Benchmark dataset.** We collect and release the first dataset of real (Community 2024)flat-pack packaging dimensions: per-box length, width, height, and weight for 202 products across 86 families. This dataset enables reproducible HFVP research with realistic item geometries rather than synthetic box generators.

**(2) LLP framework.** We propose LLP, which effectively learns the necessity of center-of-mass awareness during box placement and can robustly accommodate random changes in the spatial arrangement of items like Lego.

**(3) Results.** Five baseline heuristics tested on real-world clutter scenarios without CoM constraints 10-40% per-box stability failure in sedan trunk simulations. Training the LLP model on sequences generated with CoM constraints reduced the failure rate to 0.67% on SUV-500 and 3.3% on Sedan-500. The trained model generalised to unseen products in entirely different customer cargo configurations without additional training.

## METHODS

### 1 Dataset

This study utilises real data from IKEA furniture (flat-pack boxes) and actual vehicle trunks (sedan & SUV), which will be analysed in the following.

*1.1 Data Collection*

We modify the original TidyTuesday IKEA dataset (Community 2024) with two steps as following:

**Tier 1: Raw Data.** It contains 3,694 products scraped from IKEA's Saudi Arabia catalogue with 14 attributes including name, category, price, designer, and product dimensions and *e.t.*, and 17 categories including bookcases, sofas, and chairs and *e.t.*

**Tier 2: Filtered Data.** We filter to products with complete three-dimensional measurements between 1 and 500cm, yielding 1,899 products. The filtered dataset has slightly larger median dimensions because small accessories and incomplete entries are removed.

**Tier 3: Flat-pack Data.** Real flat-pack packaging dimensions were collected from the official website of IKEA product specification pages ("Product details" to "Packaging" section), which list the number of per-box dimensions (depth $d$, height $h$, width $w$ in cm) and weight. We systematically scraped products across the most popular IKEA families, obtaining packaging data for 202 products. In total the dataset contains 300 boxes. More details of the products from Tier 3 are listed in **Table 1**.

**TABLE 1 Comparison between the three Tiers on the products.**

| | Tier 1 | Tier 2 | Tier 3 |
|---|---|---|---|
| Total number | 3,694 | 1,899 | 202 |
| categories | 17 | 17 | 16 |
| **Dimensions** | | | |
| Depth available | 2,231 (60.4%) | 1,899 (100%) | 202 (100%) |
| Height available | 2,706 (73.3%) | 1,899 (100%) | 202 (100%) |
| Width available | 3,105 (84.1%) | 1,899 (100%) | 202 (100%) |
| Depth mean (cm) | - | 56.2 | 46 |
| Height mean (cm) | - | 113.1 | 95.2 |
| Width mean (cm) | - | 119.8 | 81.1 |
| Depth range (cm) | 1-257 | 1-257 | 1-216 |
| Height range (cm) | 1-700 | 2-301 | 4-236 |
| Width range (cm) | 1-420 | 2-420 | 5-225 |
| Volume mean (L) | - | 0.911 | 0.385 |
| Weight mean (kg) | - | 75.2 | 24.9 |

We analyse two aspects of the Tier 3 dataset: the inner ring displays six parent categories and the outer ring shows 16 subcategories with their counts (narrow categories listed in the legend), and the number of boxes per product, with around 90% of products shipped in 1 or 2 boxes. (**Figure 1**)

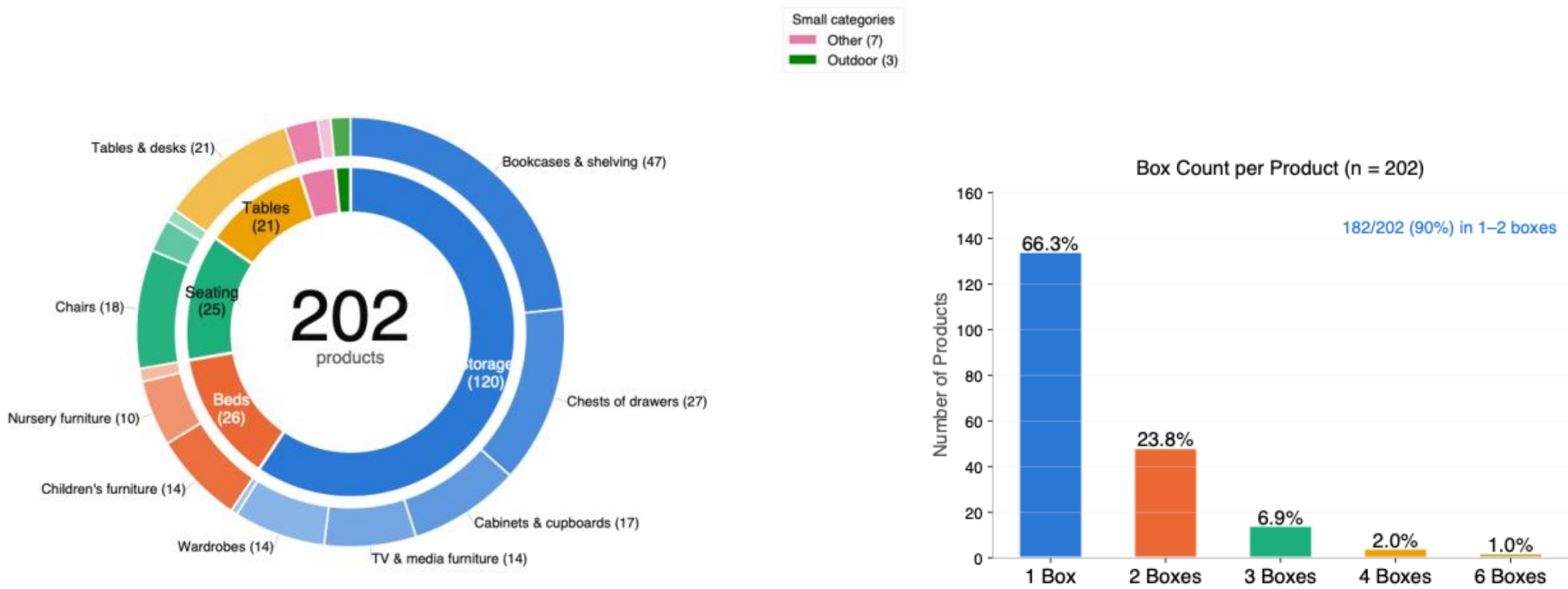

**FIGURE 1 Product category distribution (left) and box count distribution (right) for the Tier 3 dataset.**

*1.2 Assembled Products vs Flat-packs*

The distributions of assembled products against packaging boxes across the 202 products are compared on one-to-one mapping of: depth/width, height/length, width/height (log scale) and volume (log scale) (**Figure 2**). The distributions of each pair fit a Gaussian distribution. The height/length panel confirms that the longest product dimension is largely preserved in the box. The width/height panel reveals the flat-pack compression effect, where assembled product width (typically the second-largest dimension) is compressed into the thin box height dimension. The volume panel (log scale) shows packaging volume is compressed proportionally compared to assembled volume, with a median volume ratio of 4.9 times.

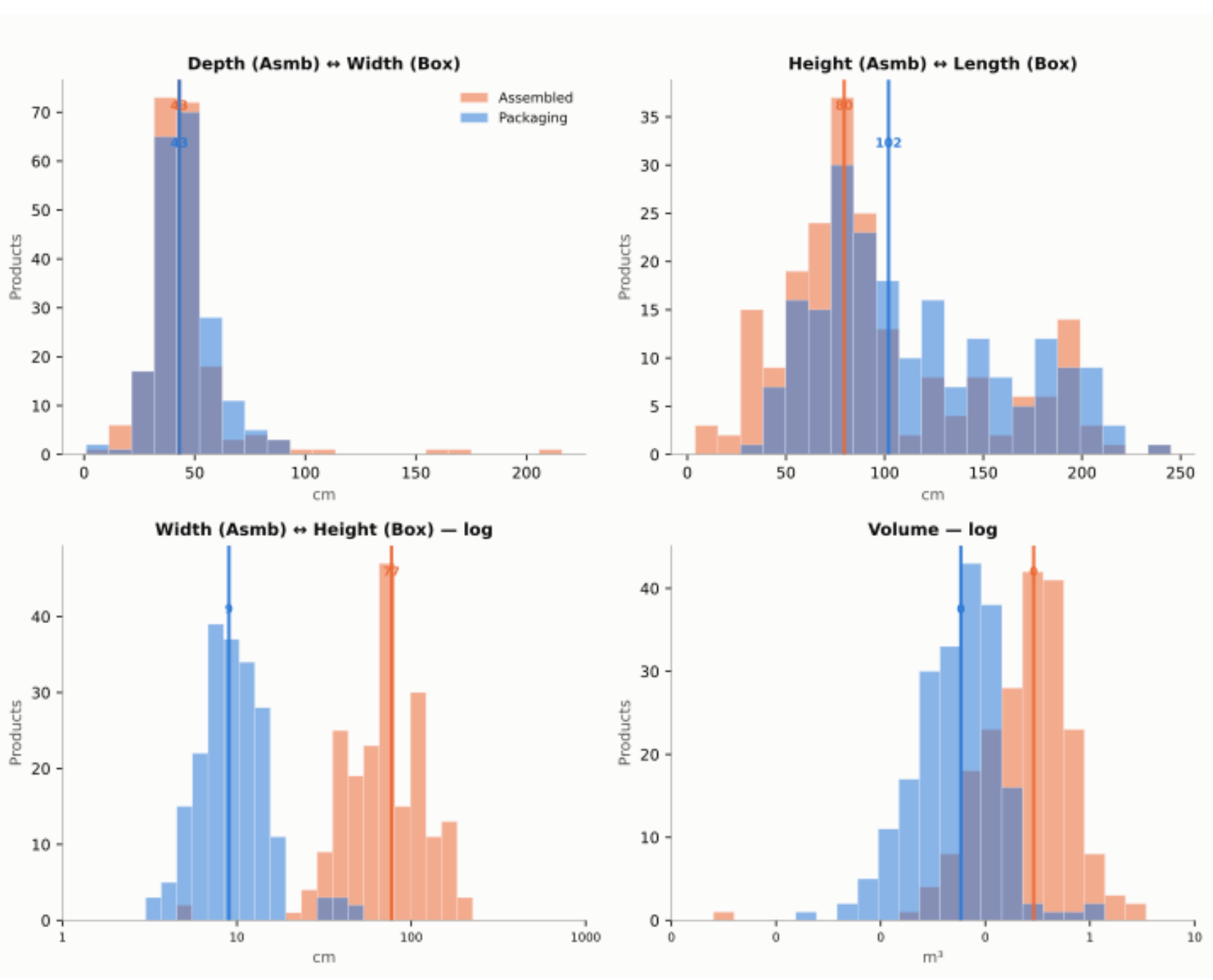

**FIGURE 2 The distributions of *assembled products* and *flat-packs* observed from four dimensions.**

We provide six IKEA product examples from the dataset spanning five categories (**Figure 3**). For each product: (1) photo (top), (2) name and category, (3) product dimensions in W×D×H (middle) with volume, (4) flat-packs with dimensions in L×W×H with weight (bottom) and weight, and (5) the volume compression ratio, which is calculated by assembled volume divided by packaging volume. Products are ordered by increasing compression, and URLs are provided for independent verification.

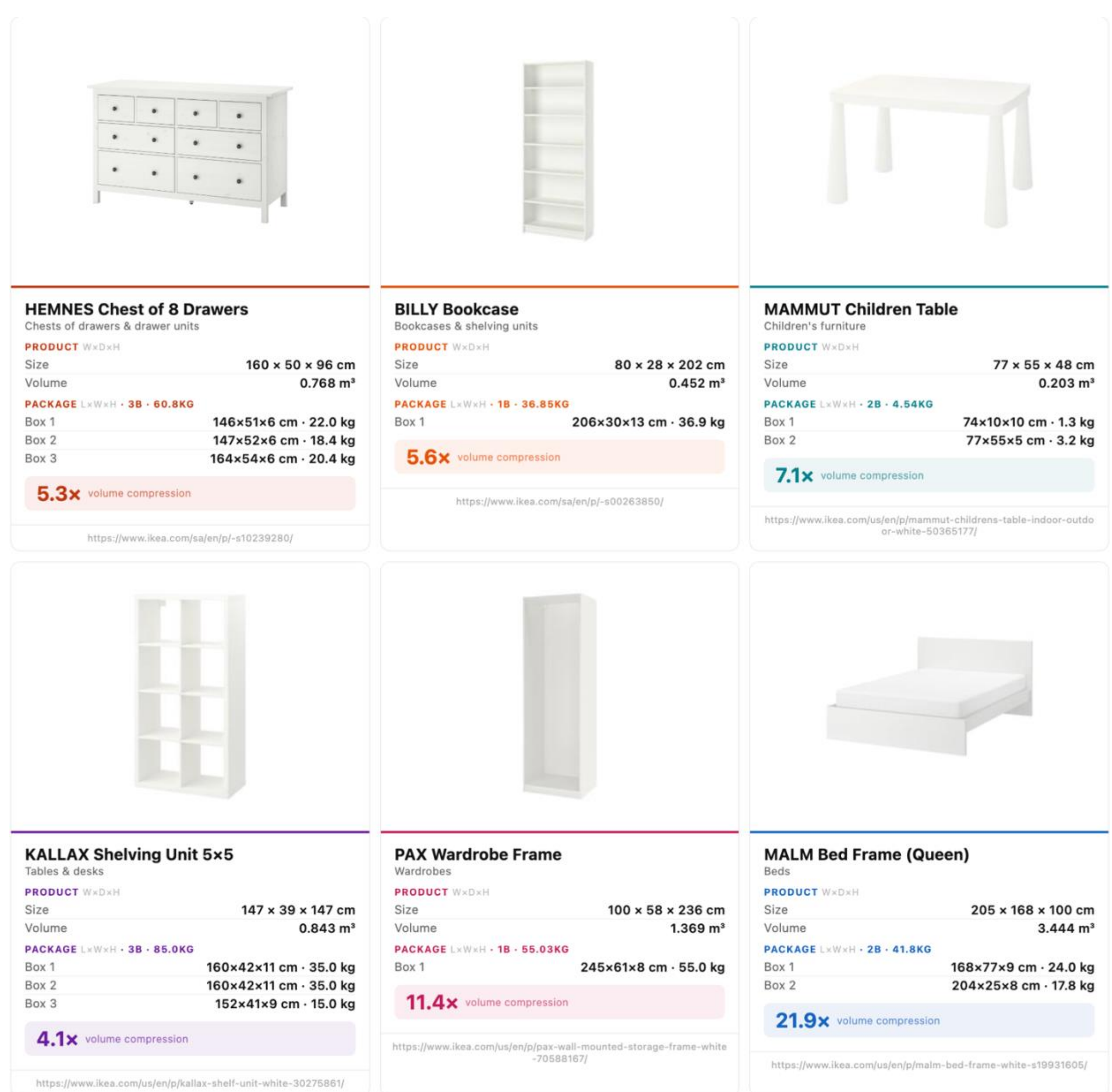

**FIGURE 3 Examples of six IKEA products from the dataset.**

*1.3 Vehicle Cargo Space*

Two vehicle trunk configurations are considered, representing common cargo spaces in passenger vehicles. The SUV configuration (based on a Chevrolet Suburban) provides a cargo area of 220×140×90 cm, while the Sedan configuration (Toyota Camry) offers a smaller 110×95×45 cm trunk. For the purpose of packing optimisation, each trunk is modelled as an empty rectangular cuboid; irregular contours, wheel-well intrusions, and seat-fold geometries are abstracted away. More details of the vehicle parameters are provided in Section **RESULTS**.

*1.4 Packing Examples*

The simulation of the items (***Section 1.2***) packing into different types of cargo (***Section 1.2***) is illustrated across four representative scenarios (**Figure 4**). Each example shows the cargo space with placed boxes rendered as colored, unplaced boxes in red, and pre-existing clutter in grey. The MAMMUT children's table (**Figure 3**) is highlighted with leader lines to show how a single product can appear in multiple scenarios.

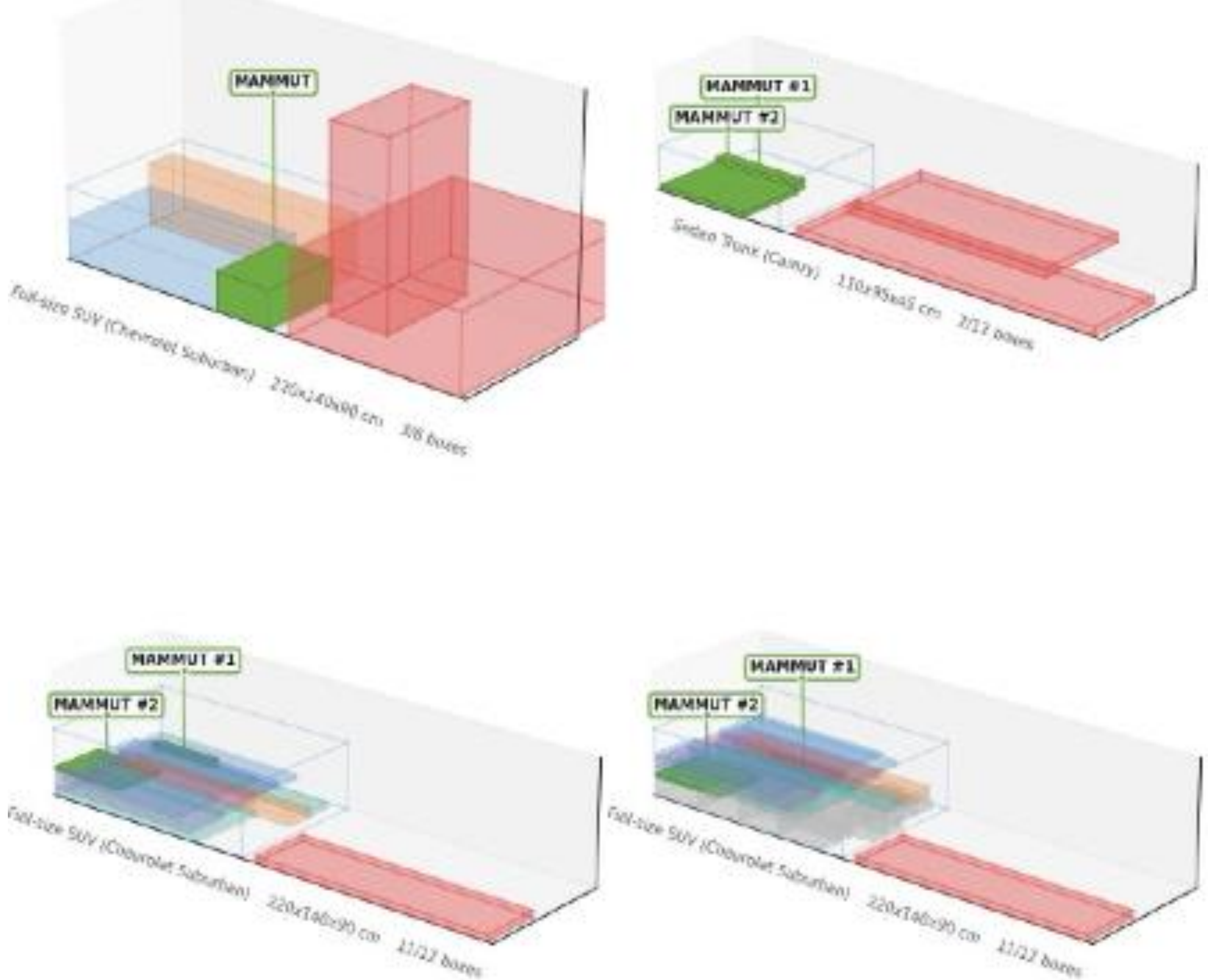

**FIGURE 4 Four illustrative HFVP scenarios. Unplaced boxes shown in red wireframe; MAMMUT items highlighted with leader labels.**

**(a) SUV (Assembled furniture).** Only 3/6 items are fit into the SUV because assembled volumes are much larger than their flat-pack equivalents, demonstrating why flat-pack packaging is essential to personal-vehicle transport.

**(b) Sedan (flat packs).** A sedan (110×95×45 cm) accommodates only 2/12 flat-pack boxes with overcapacity. Long items such as PAX (245 cm) and BILLY (206 cm) cannot fit in any orientation. This illustrates that the depth constraint on vehicle length is the binding limitation for small vehicles.

**(c) SUV** **(flat packs)**. The longer cargo floor directly enables more placement, where 11 boxes are placed, with only the PAX (245 cm) remaining unplaced. *In practice, this item can be placed into the trunk at an angled position, from a manual handling perspective.*

**(d) SUV with clutter** **(flat packs)**. The same vehicle is pre-loaded with 25 random-sized items occupying around 18% of the cargo volume, where *clutter is put into corners to replicate actual loading environments*. The remaining irregular space provides unstable spaces for the item packing.

## 2 Algorithms

### *2.1 Problem Definition*

Let $F = \{f_1, \cdots, f_N\}$ be a catalogue of furniture items where each $f_i$ is characterised by assembled dimensions $(d_i, h_i, w_i)$ as *(depth, height, width)* in cm. To simply this task, each box corresponds to one IKEA product. Let $B = \{b_1, \cdots, b_N\}$ be a set of flat-pack boxes where each $b_i$ has packaging dimensions $(l_i, w_i, h_i)$ as *(length, width, height)* in cm.

A packing process is defined by a subset of boxes $O \subseteq B$ representing one customer order, and a vehicle cargo space $\mathrm{C}$ with dimensions $(\mathrm{D}, \mathrm{W}, \mathrm{H})$ representing (Depth, Width, Height) in cm. The cargo space may contain pre-existing clutter items that occupy fixed positions and cannot be moved.

**Goal:** To define a placement policy $\pi$ that operates on the union of all flat-pack boxes $O \subseteq B$ *and* maps the entire collection to an **ordered sequence** of placement within the target container $C$.

$$\pi\!: \bigcup_{f_i \in O_k} \mathcal{B}_i \rightarrow \{(pose_1, \theta_1), (pose_2, \theta_2), \ldots, (pose_n, \theta_n)\} \subset \mathcal{C}_j$$

where each box $\mathrm{b_i}$ is put with corner position $(\mathrm{x_i}, \mathrm{y_i}, \mathrm{z_i})$ and an orientation $\theta_i \in \{0, \ldots, 5\}$ (six axis-aligned rotations), maximising the number of placed boxes while satisfying three constraints

**Constraints**:

**(a) Non-overlap.** Oriented bounding boxes must not intersect each other or pre-existing clutter.

**(b) Containment.** Each box must lie entirely within the cargo space.

**(c) Per-box support stability.** Each box placed above the floor ($z > 0$) must have its Center-of-Mass within the convex hull of its support points, with an anti-tipping margin proportional to box height (as discussed in the stability section).

### *2.3 Center-of-Mass*

Observations from the visualisation result (**Figure 4**) show that, in real-world loading scenarios, flat-pack boxes are typically long and narrow, and the presence of clutter further enhances the balance problem. The stability constraint ensures that each placed box remains stable without tipping or sliding off its support and serves as a placement-level physical feasibility constraint.

After placing a flat-pack box $b_i$ at $(x_i, y_i, z_i)$ with $(l_i, w_i, h_i)$, its support comes from occupied cells directly beneath it. Let $\mathcal{G}$ be the voxel occupancy grid after previous placements. Let $\mathrm{u}, \mathrm{v}$ be integer grid indices (rows and columns). The support contact points are the centers of occupied cells on the box footprint:

$$\mathcal{S}(b_i) = \{\, (u + 0.5,\ v + 0.5) \mid x_i \le u < x_i + l_i, y_i \le v < y_i + w_i, \mathcal{G}[u, v, z_i - 1] = 1 \,\}$$

Ithe box is on the floor and unconditionally stable when $\mathrm{z_i} = 0$.

The center of mass projects to the geometric center $(c_{x,i},\ c_{y,i}) = (x_i + \mathrm{l}_i/2, y_i + w_i/2)$. Four conditions determine stability:

- **(a) Support existence.** $|\mathcal{S}(b_i)| = 0$ is rejected unconditionally.
- **(b) Degenerate support.** $|\mathcal{S}(b_i)| \in \{1,2\}$ (point/line contact) is only accepted for boxes shorter than 5 cm; a tolerance of $0.05 \cdot \max(\mathrm{l}_i, w_i)$ governs CoM deviation from the contact point(s). All taller boxes on 1–2 point supports are unconditionally rejected.
- **(c) Convex hull containment.** With $|\mathcal{S}(b_i)| \ge 3$, let $\mathcal{H}$ be the convex hull of $\mathcal{S}(b_i)$. The CoM must lie within $\mathcal{H}$: $(c_{x,i}, c_{y,i}) \in interior(\mathcal{H})$

**(d) Anti-tipping margin.** A tall box on a narrow support will tip under lateral force even with CoM inside $\mathcal{H}$. Let $d_{min}$ be the shortest Euclidean distance from $\left(c_{x,i}, c_{y,i}\right)$ to any edge of $\mathcal{H}$. We require $d_{min} \geq \alpha \cdot h_i$, where $\alpha = 0.15$ a conservative ratio that ensures a sufficient restoring moment against lateral forces.

The algorithm integrates these checks into placement on each step and the occupancy grid is updated after. All four conditions are evaluated, and the placement is either committed or rolled back. This guarantees that every box in the final packing satisfies per-box stability, at the cost of rejecting some geometrically feasible but physically unstable positions.

*2.3 Baseline methods*

We use five strategies as baseline methods for 3D packing from the bottom-left (BLF) packing method. Let $v_i = \ l_i \cdot w_i \cdot h_i$ denote the volume of $i$th box, $m_i$ as weight, and $\lambda_i = max(l_i, w_i, h_i)$ as the longest dimension.

**(a) Large-First.** Sort by volume descending of $v_i$. Bulky items claim floor positions first, leaving gaps for smaller boxes.

**(b) Heavy-First.** Sort by weight descending of $m_i$. Heavy items placed low can help to reduce the overall centre of mass.

**(c) Small-First.** Sort by volume ascending of $v_i$. Small boxes fill interstitial gaps before large items are placed.

**(d) Longest-First.** Sort by the longest dimension descending of $\lambda_i$. The longest boxes, which are most constrained by vehicle depth, are placed first.

**(e) Random-Best.** Generate random permutations, keep the ordering that yields the most placed boxes under the target placer.

We test the five methods with the example dataset (shown in **FIGURE 3**). Using CoM as a hard gate during placement creates a stark divergence. Large-First (a), Longest-First (d), and Random-Best (e) all achieve full occupancy with zero stability failures, demonstrating that CoM-compliant orderings exist for this scenario. Heavy-First (b) places 10/11, where a single MALM bed frame box fails to find a stable position under the weight-prioritised order. Small-First (c) collapses to 2/11 because the first two small boxes (MAMMUT Children 1&2) consume the available floor area with narrow support footprints, blocking all nine subsequent boxes. The step-by-step rendering process is shown below (Figures x) for result visualisation.

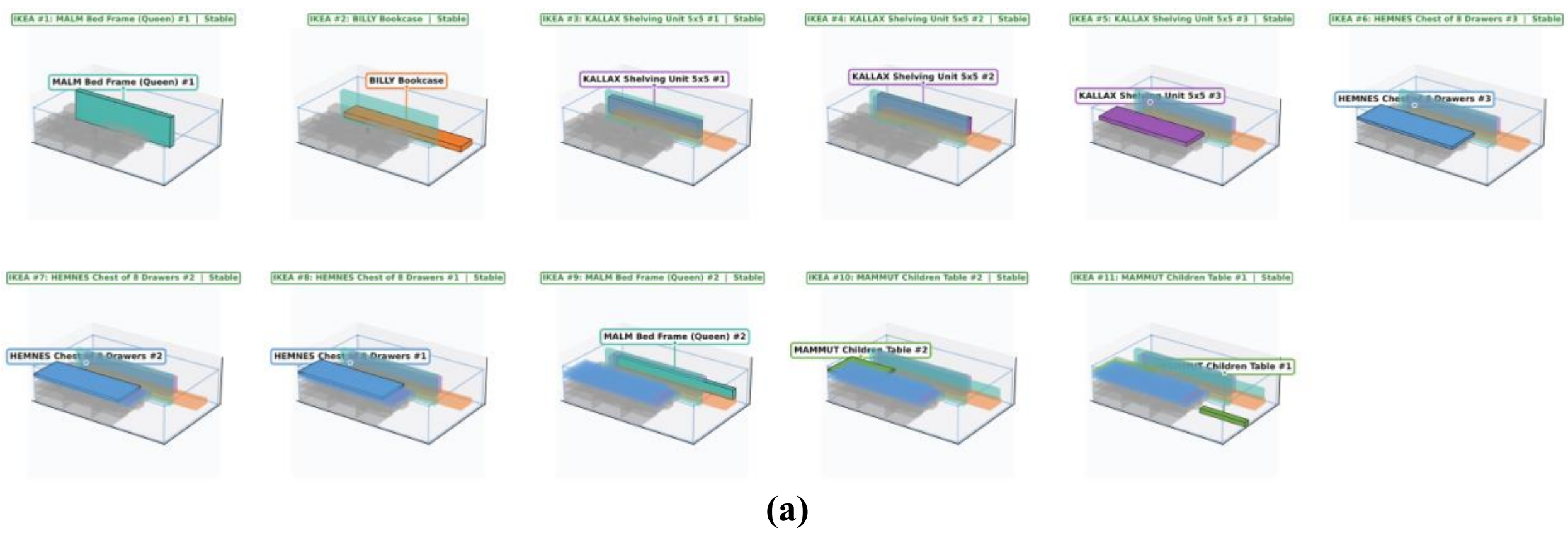

**(a)**

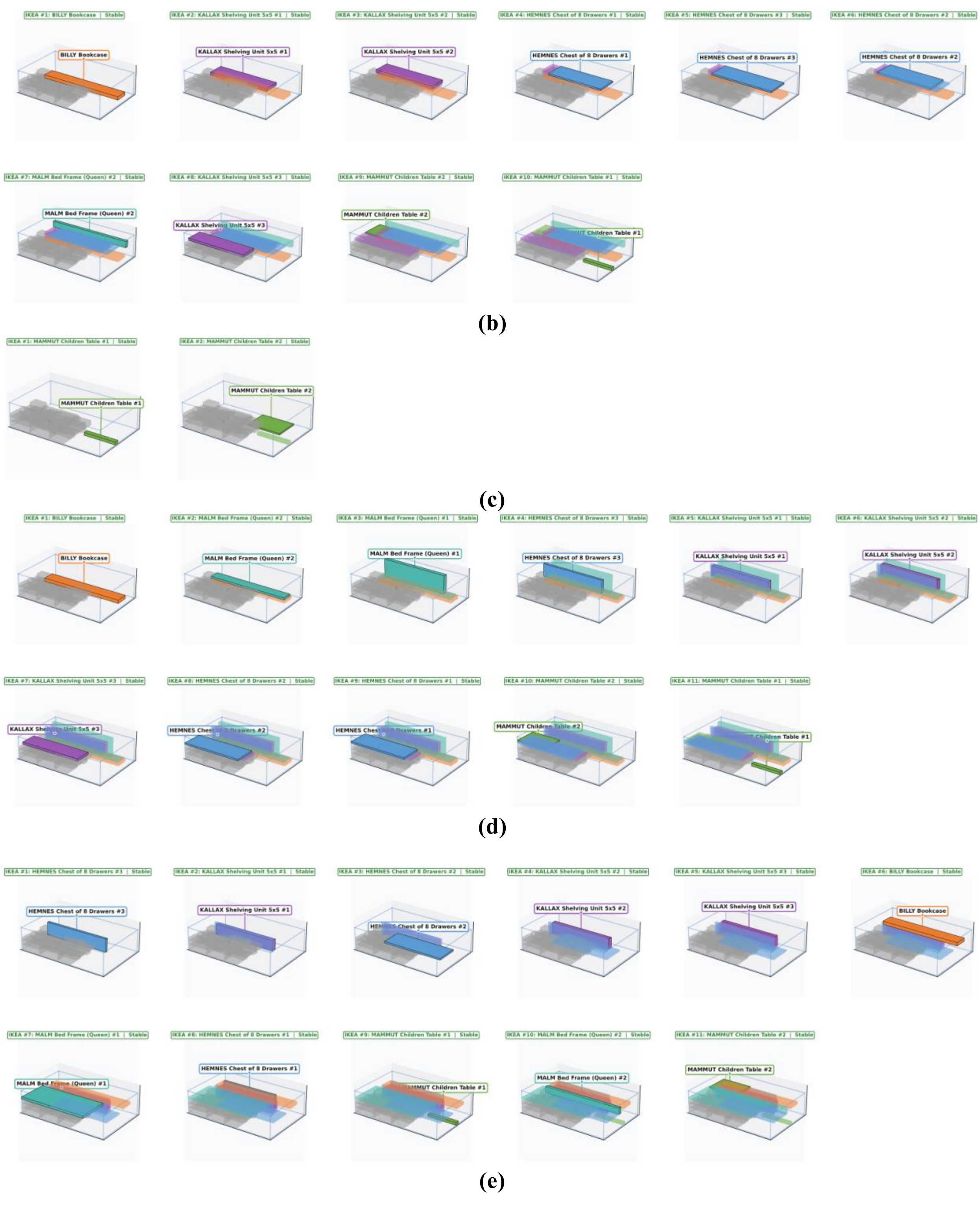

**FIGURE 5 Step-by-step visualisation results of the packing process using the five baseline methods using the example flat-pack boxings. The same item box remains the same colour.**

*2.3 Lego-Language Packing*

We adopt a Transformer-based large language model, LLaMA-3.2-1B-Instruct, and fine-tune it to generate packing sequences autoregressively. The overview of the three-stage pipeline is shown in **Figure 6**. **(a) Tokenisation:** The description and purchased product information are encoded into a token sequence as input to the model. **(b) Autoregressive Training:** The tokenised sequences, with input box order variants, are fed to the language model for autoregressive training. The model also generates a sequence output, where each new token depends on all tokens that came before it. **(c) Inference:** The user provides product names in arbitrary order, and they are converted to IDs via catalogue lookup and formatted into the prompt as input. The model outputs a new sequence of the item IDs with left-corner position and rotations.

.

*Tokenisation and Vocabulary Alignment*

Let $\mathcal{P}$ denote the prompt tokens encoding the task specification (vehicle, catalog, grid, format) as natural language. Let $\mathcal{S}$ be the SEP token, and $\mathcal{E}$ the EOS token. Let $\mathcal{I} = \{ID_1, \dots, ID_N\}$ be the set of box ID tokens to be packed. Let $\mathcal{O} = (\mathrm{o}_1, \dots, \mathrm{o}_\mathrm{N})$ be the canonical output sequence, where each $o_i = (ID_i, x_i, y_i, z_i, \theta_i)$ is a 5-token placement:

- **(a) Prompt.** Encodes the task specification information such as vehicle dimensions, catalogue of available box IDs, grid resolution, and output format.
- **(b) Input IDs.** Lists the IDs of all boxes that need to be packed; shuffled during training to produce diverse variants, provided in arbitrary user order during inference.
- **(c) SEP.** A dedicated separator token that marks the boundary between the input and output.
- **(d) Output placements.** The target placement sequence provides each step of the $o_i = (ID_i, x_i, y_i, z_i, \theta_i)$
- **(e) EOS.** Signals the end of the output sequence

The input or the training process consists of the concat:

$$\mathcal{T}t = \mathcal{P} \circ \text{shuffle}(\mathcal{I}) \circ \mathcal{S} \circ \mathcal{O}$$

The input or the inference process consists of the concat:

$$\mathcal{T}i = \mathcal{P} \circ \mathcal{I} \circ \mathcal{S}$$

The output or the train/inference process consists of the concat:

$$\mathcal{Y} = \mathcal{O} \circ \mathcal{S}$$

*Autoregressive Training*

Let $\mathcal{T} = (\mathrm{t}_1, \dots, \mathrm{t}_\mathrm{M})$ be the full training sequence. The model reads all preceding tokens $\mathrm{t}_{<\mathrm{k}}$ and predicts the next token $\mathrm{t}_\mathrm{k}$. Only output positions contribute to the loss:

$$\mathcal{L} = -\sum_{k \in \mathcal{K}_{out}} log\, P(t_k \mid t_1, \dots, t_{k-1}), \qquad \mathcal{K}_{out} = \{k \mid t_k \in \mathcal{O} \cup \{\mathcal{E}\}\}$$

*Inference*

At inference, the user supplies an unsorted list of product names (e.g., KYRRE, BILLY, TROFAST, MALM, KALLAX). The system converts each name to its ID via the catalogue lookup and constructs the prompt with vehicle specs, catalogue listing, and format instructions. The model then generates the placement sequence token by token, where $\widehat{o_t}$ present the prediction of $t$-th step:

$$\widehat{o_t} \sim P(\cdot \mid \mathcal{P}, \mathcal{S}, \mathcal{I}_{\mathrm{ue}}, \widehat{o_1}, \dots, \widehat{o_{t-1}}), \qquad t = 1, \dots, N$$

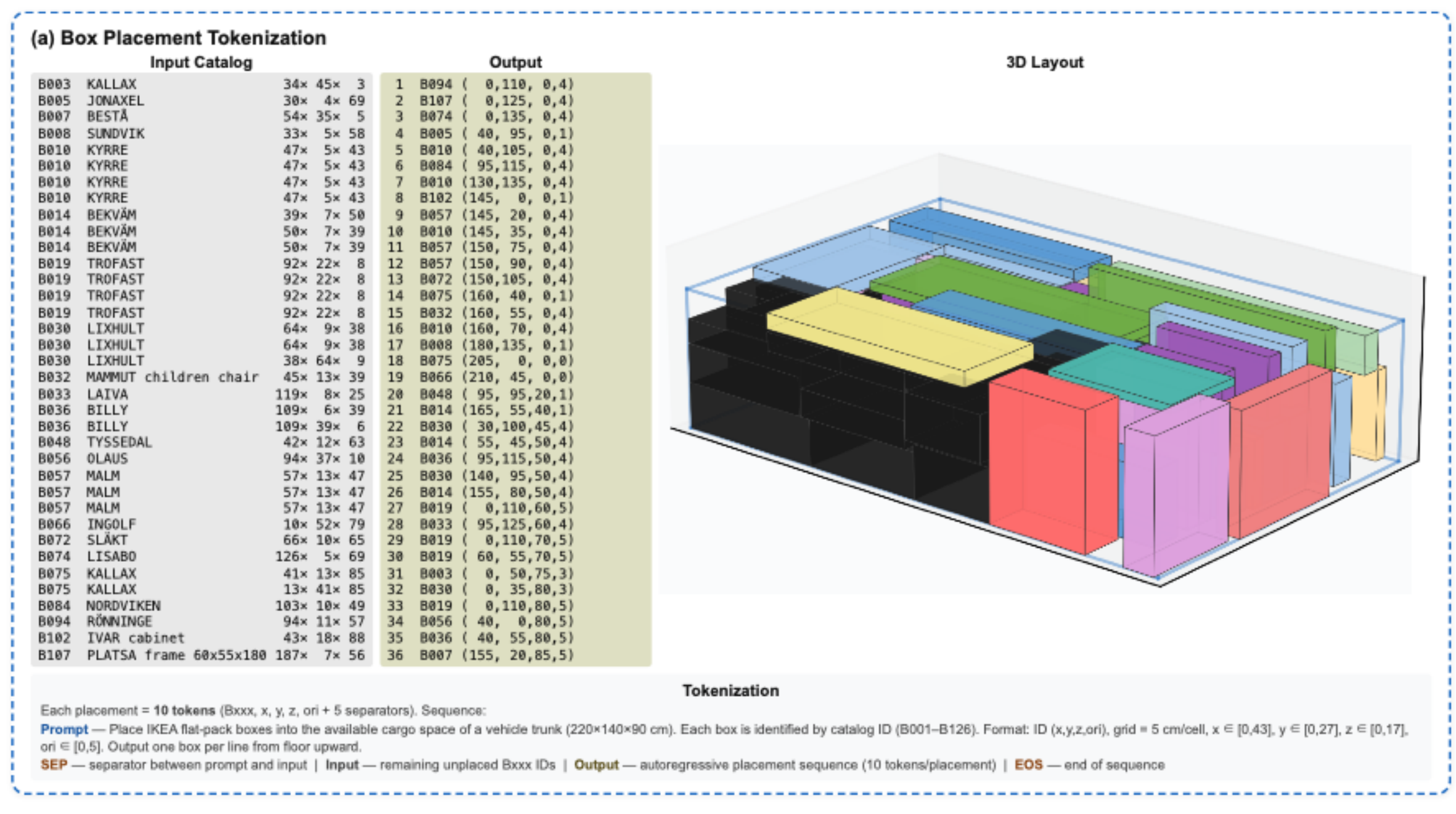

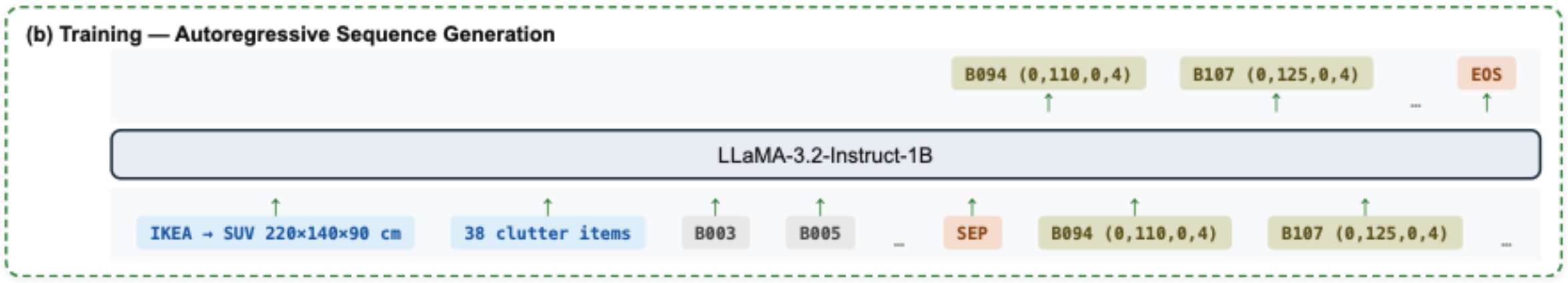

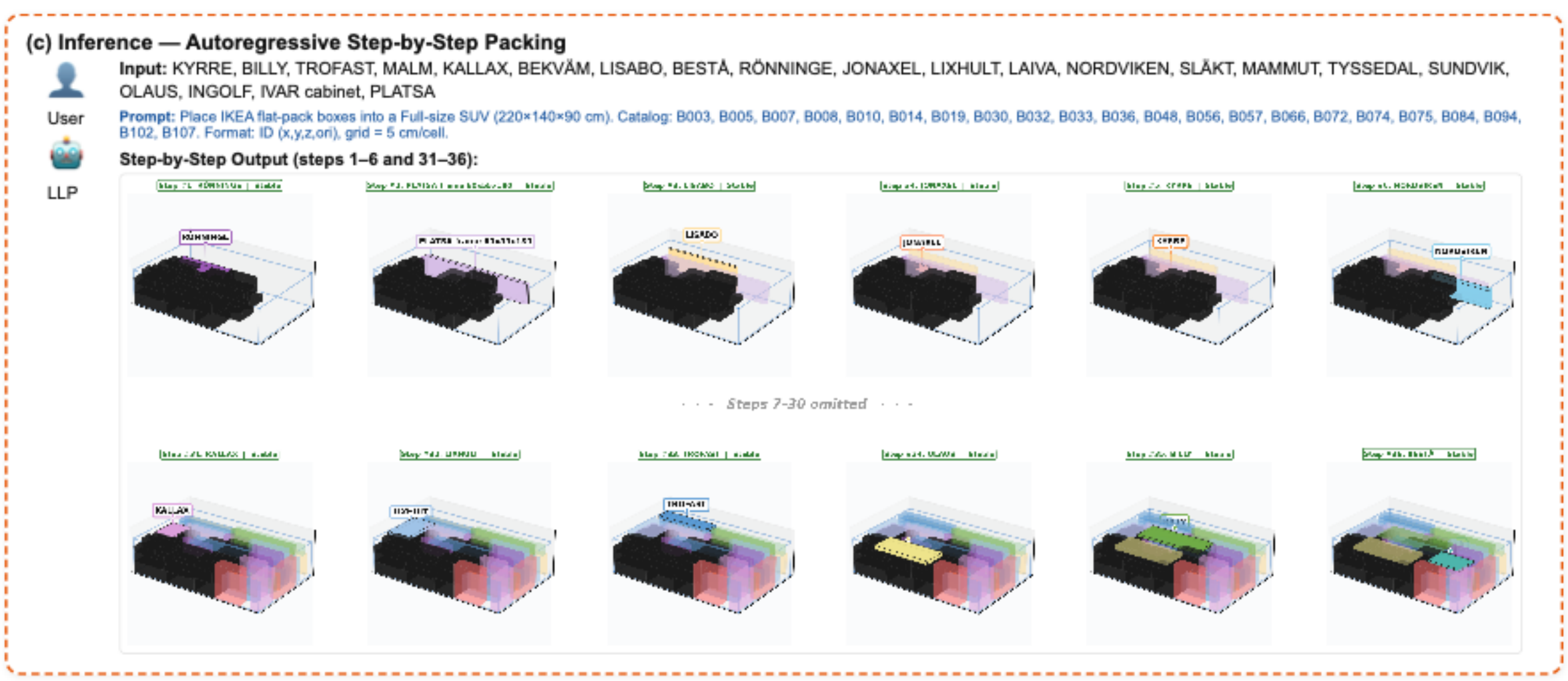

**FIGURE 6 Our Lego-Language Packing framework.**

*LEGO Formate Data Generation*

A real vehicle trunk is never a clean rectangular volume with strollers, groceries, and sports equipment already occupying irregular regions. **(1) Choosing flat-boxes:** We reverse the LEGO disassembly process where a completed LEGO model can be taken apart brick by brick (each removal exposes the structure beneath), a packed vehicle can be unpacked by removing boxes one at a time from the top down, revealing the supporting layers. **(2) Generating steps:** Taking all blocks, starting from the cluttered floor, boxes are placed one at a time into the
remaining irregular gaps, bottom-up. Each box sits at the top of clutter or previously placed boxes. Now we can generated infinity number of training datasets with **different clutter and product catalogues**.

**(a) Voxel representation:** Spatial reasoning requires discretisation: a 5~cm voxel grid enables fast placement search because when it comes to large space, the searching time will be costly. Checking whether an oriented box fits at a candidate position requires only constant time per candidate under a voxel representation.

**(b) CoM-gated.** Every box in the training data must satisfy per-box CoM stability, where the model can be trained to learn. The generator enforces this adaptively: a fast 5~cm pre-filter rejects candidates with support ratio below 35%, then a precise 1~cm convex hull check evaluates the four stability conditions (support existence, degenerate contact, hull containment, anti-tipping margin). Only candidates passing both are committed.

**(c) Randomised box inventory.** Each scenario draws a random subset of 50 box types from the 126-type catalogue where the eval sequence is unknown during the training process. To enhance the randomness of the dataset to simulate human behaviour, we randomly limit the number of every object, where the large objects are limited to the number from 1-3 and small objects can be up to 10 to fit the corner of the cargo space.

# RESULTS

## 1 Experience Settings

### *1.1 Dataset Settings*

**TABLE 2 Dataset for training and evaluation**

| | SUV-500 | Sedan-500 |
|---|---|---|
| Vehicle | Chevrolet Suburban | Toyota Camry |
| Cargo dims (cm) | 220x140x90 | 110x95x45 |
| Clutter | 38 items (34% vol) | None |
| Scenarios | 500 | 500 |
| boxes per scenario | 50.2 | 25.7 |
| boxes types per scenario | 10.2 | 5.8 |
| Unique IDs covered | 93/126 (73.8%) | 45/126 (35.7) |
| Gen. time per scenario (s) | 45 | 9 |
| Total gen. time (h) | 6.4 | 1.3 |
| Volume utilisation (%) | 54.3 | 46.3 |

We generate two datasets on different vehicle types (**Table 2**). Each dataset contains 500 scenarios generated for our LLP framework, split 80/20 at the scenario level (400 train, 100 eval) with 5 shuffle variants per scenario, yielding 2,000 training and 500 evaluation samples. The token count is around 221 (SUV) and around 147 (Sedan). CoM compliance is 100% by construction. The main reason for the time cost is from the dominant cost of the CoM convex hull check with an irregular surface. The 5cm pre-filter eliminates approximately 97% of candidates cheaply, but the surviving 43,000 (SUV) vs. 12,000 (Sedan) candidates per scenario each require a full support-point convex hull computation at $\mathcal{O}(\mathrm{n}\log\mathrm{n})$ operations over the 1cm footprint cells.

### *1.2 Training Settings*

The training model is LLaMA-3.2-1B-Instruct (1B parameters, decoder-only Transformer, 32 attention heads, 16 layers, hidden size 2048), fine-tuned with LoRA (rank r = 32, scaling $\alpha = 16$, dropout 0.05) . The tokeniser is extended with 126 special tokens from B001 to B126. Prompt and input tokens are masked. Training uses the TRL with AdamW (learning rate $\text{lr} = 2 \times 10^{-4}$, cosine schedule, 20 warmup steps), 5 epochs, batch size 1 with gradient accumulation over 8 steps (effective batch 8), and max sequence length 2,048 tokens. Training completes in around 14min on a single NVIDIA 4090-24GB GPU for the full 500 scenarios.

**2 Comparison**

Our fine-tuned LLaMA-3.2-1B model achieves a dramatic reduction in centre-of-mass instability compared to all BLF heightmap-based baselines. On the SUV-500 configuration, our model attains a CoM failure rate of just 0.78%, representing a huge improvement over all the BLF. On the Sedan configuration, our model achieves a 3.3% failure rate, reducing instability by a factor of 3.4 relative to the strongest BLF baseline (11.3%). This improvement proves the model's ability to find the balanced packing process, effectively learning a physically grounded placement policy. (**Table 3**)

**TABLE 3 Comparison on the evaluation set of the SUV-500 dataset**

| | No. Placed | No. CoM Fail | Fail Rate % |
|---|---|---|---|
| GT | 5,052 | 0 | 0.0 |
| Large First (a) | 5,012 | 1,695 | 33.8 |
| Heavy First (a) | 5,011 | 1,714 | 34.2 |
| Small First (a) | 5,016 | 946 | 18.9 |
| Longest First (a) | 4,977 | 1,350 | 27.1 |
| Random Best (a) | 5,039 | 1,265 | 25.1 |
| LLP (ours) | 4,911 | 33 | 0.67 |

Although our model is much better at stability, it packs fewer boxes than BLF baselines. This gap comes from a key limitation of the language model approach for spatial tasks, where the LLMs have illusions to fully place all boxes one by one. BLF puts each box one after another, which will scan all the items without missing. A good way to keep the stability advantage while fixing the illusion problem is to combine them. For example, constrained decoding, collision-checked rejection sampling, or a hybrid can be added to runs after the LLM. (**Table 4**)

**TABLE 4 Comparison on the evaluation set of Sedan-500 dataset**

| | No. Placed | No. CoM Fail | Fail Rate % |
|---|---|---|---|
| GT | 2,539 | 0 | 0.0 |
| Large First (a) | 2,468 | 425 | 17.2 |
| Heavy First (a) | 2,467 | 423 | 17.1 |
| Small First (a) | 2,417 | 273 | 11.3 |
| Longest First (a) | 2,477 | 407 | 16.4 |
| Random Best (a) | 2,509 | 336 | 13.4 |
| LLP (ours) | 2,397 | 78 | 3.3 |

We tested the effect on the number of training samples. The ablation study that the learned CoM policy is highly sample-efficient without a huge amount of data. Training on as few as 100 scenarios (500 samples) already drops the CoM failure rate below 2%. This saturation behaviour indicates the CoM can learn the balance information with limited data, which will not cost much when implementing into a new environment. (**Table 5**)

**TABLE 5 Comparison on the number of training scenarios of Sedan-500 dataset**

| | No. Placed | No. CoM Fail | Fail Rate % |
|---|---|---|---|
| 100 | 2,274 | 42 | 0.18 |
| 200 | 2,283 | 66 | 0.28 |
| 300 | 2,318 | 50 | 0.21 |
| 400 | 2,350 | 78 | 0.33 |
| 500 (ours) | 2,397 | 78 | 0.32 |

## 3 Visualisation

We visualise the step-by-step packing results on both our proposed LLP method and the baseline small-first method. It is observed that our method initially chooses the large items to occupy the void on the right side, followed by filling the remaining large-area spaces at the front. Most boxes are in stand position, exhibiting a high orientation variation. The small-first baseline, while also covering the right-side area but filling with small items, results in a total of 7 unbalanced locations and underperforms compared to our trained model. (**Figure 7**)

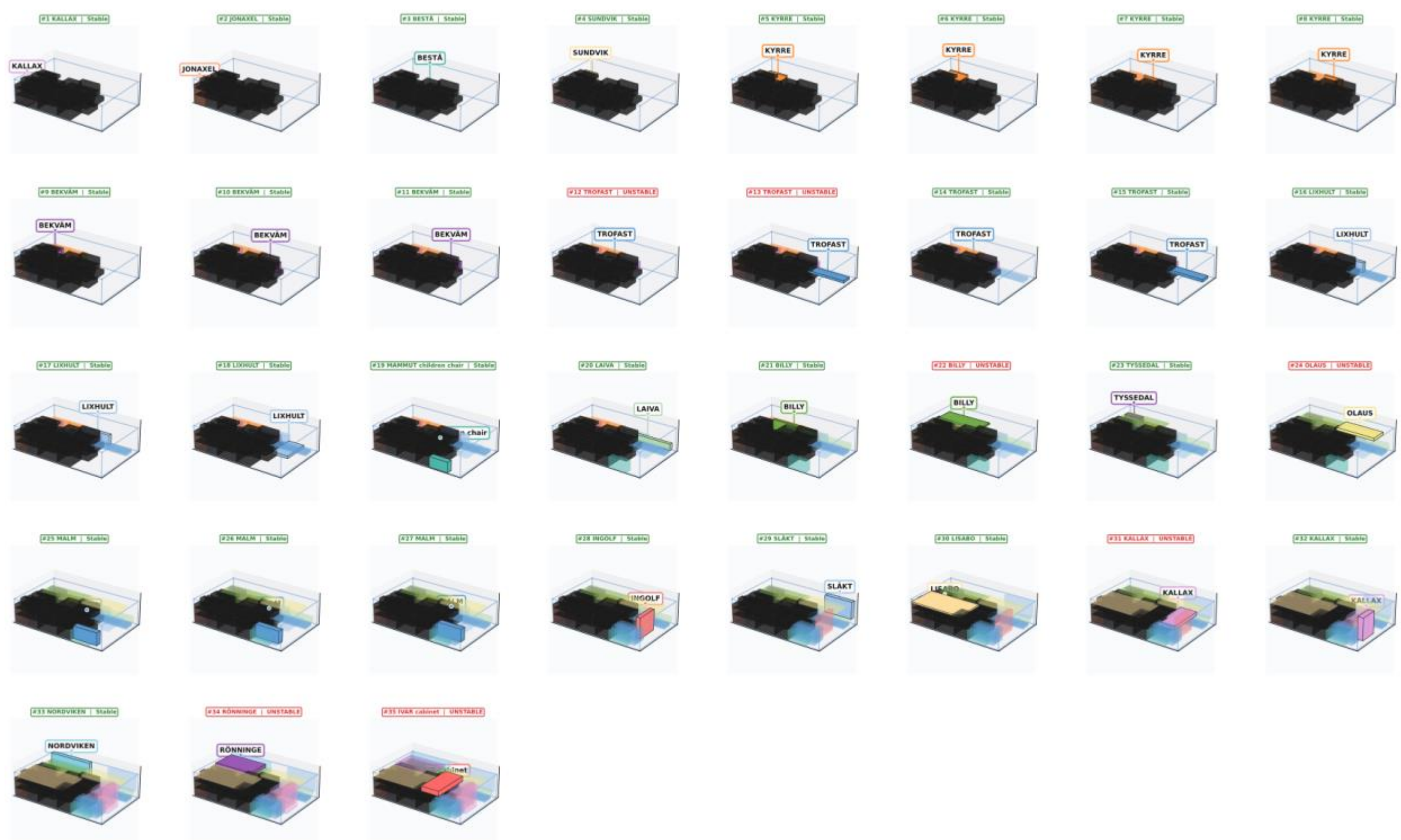

**FIGURE 7 One packing example from our method vs small first on the SUV-500 dataset.**

## DISCUSSION

Real-world furniture loading differs from standard bin packing in three ways: (1) the cargo space is partially occupied by random clutter, (2) the boxes are heterogeneous and slender, and (3) the flat-pack boxes are thinner and taller. These challenges make per-box CoM stability the critical constraint. Our model is trained on CoM-gated sequences packing data from real IKEA flat-packages. After fine-tuning, it learns stability information and predicts each candidate position, generalising across different box inventories and clutter arrangements without re-training.

Limitations: The current settings simplify the constraints, with only balance considered for packing. We assumed boxes are always placed straight (aligned with the axes). In reality, people often tilt them to put the long boxes into the trunk. Our method only allows six fixed orientations, which doesn't cover this operation. We also ignored some real-world rules that some boxes can't have anything stacked on top.

Future work. Our token-based approach is promising for real-world use, because we can add practical constraints (like crush-sensitive boxes, unloading order, and axle weight limits) without changing the model itself. We just describe them as structured tokens in the input. Even better, if we train on real loading data from furniture logistics where human packers already follow these rules unconsciously, the model can learn multi-constraint packing strategies by imitating humans, instead of relying on hand-coded rules. This progression, from synthetic single-constraint data to real multi-constraint supervision, is the natural next step for the large language model.

## CONCLUSIONS

This paper addressed the problem of packing heterogeneous flat-pack furniture into a partially occupied vehicle under per-box stability constraints. We proposed the LEGO-Language Packing (LLP) framework, which reframes spatial placement as autoregressive token generation. Three contributions were made. First, a data generation pipeline produces CoM-compliant packing sequences from a dual-resolution voxel grid with randomised box inventories, yielding 500 scenarios across 126 IKEA product types. Second, a token placement encoding enables a standard causal language model to predict box positions, orientations, and stability outcomes from the tokenised prompt and input inventory. Third, a benchmark of five sorting strategies reveals that sort order dominates placement success when CoM is enforced as a hard gate better than previous methods. The autoregressive token paradigm opens a path toward training larger models on real loading data with richer operational constraints.

**ACKNOWLEDGMENTS**

We use the AI/LLM models for two purpose in this paper. Firstly, we used DeepSeek for grammar checking and language polishing, after all revisions manually are verified by the authors finally. Secondly, we use  the Claude add-on in GitHub Copilot to generate and run background scripts for research automation running, and we validated all outputs and codes from cross verification from different setting methods.

**AUTHOR CONTRIBUTIONS**

The authors confirm contribution to the paper as follows: study conception and design: H. Li, Y. You; data collection: H. Li; analysis and interpretation of results: H. Li; draft manuscript preparation: H. Li, Y. You. All authors reviewed the results and approved the final version of the manuscript.

**DECLARATION OF CONFLICTING INTERESTS**

The authors declared no potential conflicts of interest with respect to the research, authorship, and/or publication of this article.

**FUNDING**

The authors disclosed receipt of the following financial support for the research, authorship, and/or publication of this article: This research was supported by The Hong Kong Polytechnic University.